\documentclass[a4paper]{article}
\usepackage{iwslt15,amssymb,amsmath,epsfig}
\usepackage{hyperref}
\usepackage{float} 
\def\reg{{\rm\ooalign{\hfil
     \raise.07ex\hbox{\scriptsize R}\hfil\crcr\mathhexbox20D}}}

\title{Data Selection with Cluster-Based Language Difference Models\\ and Cynical Selection} 

\name{\em Luc\'ia Santamar\'ia, Amittai Axelrod}
\address{Amazon.com \\  {\small \tt \{lucsan,amittai\}@amazon.com} }

\begin{document}
\maketitle
\begin{abstract}

We present and apply two methods for addressing the problem of selecting relevant training
data out of a general pool for use in tasks such as machine translation. Building on existing work on class-based language difference models~\cite{AxelrodVyasMC:2015}, we first introduce a cluster-based method that uses Brown clusters to condense the vocabulary of the corpora. Secondly, we implement the cynical data selection method~\cite{Axelrod:2017}, which incrementally constructs a training corpus to efficiently model the task corpus. Both the cluster-based and the cynical data selection approaches are used for the first time within a machine translation system, and we perform a head-to-head comparison.

 Our intrinsic evaluations show that both new methods outperform the standard Moore-Lewis approach (cross-entropy difference), in terms of better perplexity and OOV rates on in-domain data. 
 The cynical approach converges much quicker, covering nearly all of the in-domain vocabulary with 84\% less data than the other methods.

Furthermore, the new approaches can be used to select machine translation training data for training better systems.
Our results confirm that class-based selection using Brown clusters is a viable alternative to POS-based class-based methods, and removes the reliance on a part-of-speech tagger. Additionally, we are able to validate the recently proposed cynical data selection method, showing that its performance in SMT models surpasses that of traditional cross-entropy difference methods and more closely matches the sentence length of the task corpus.
\end{abstract}

\section{Data Selection, Previously}

\subsection{Moore-Lewis Data Selection}

The standard data selection method of Moore and Lewis~\cite{MooreLewis:2010} uses cross-entropy difference as the similarity metric to estimate the relevance of each sentence in the general pool corpus. This method takes advantage of the presumed mismatch between the pool data and the task domain. It first trains an in-domain language model (LM) on the task data, and then trains another LM on the full pool of general data. The average per-word perplexity of each sentence in the pool data is computed relative to each of these models. The cross-entropy $H_{lm}(s)$ of a sentence $s$, according to language model $lm$, is the log of the perplexity of the language model on that sentence. The cross-entropy difference score of~\cite{MooreLewis:2010}  is:

\begin{equation*}
 \label{eqn-xediff}
    H_{LM_{TASK}}(s) - H_{LM_{POOL}}(s).
\end{equation*}

Sentences that are most like the task data, and most unlike an average sentence in the full pool will have lower cross-entropy difference scores. A modification of this method, the bilingual Moore-Lewis criterion~\cite{AxelrodHeG:2011} used for selecting bilingual data for machine translation. This is a simple extension, combining the cross-entropy difference scores from each side of the corpus; i.e. for sentence pair $\langle s_1, s_2\rangle$
 \begin{eqnarray*}
 \label{eqn-bilingual-xediff}
       & \left( H_{LM_{TASK_1}}(s_1) - H_{LM_{POOL_1}}(s_1) \right) \\
    + & \left( H_{LM_{TASK_2}}(s_2) - H_{LM_{POOL_2}}(s_2) \right).
\end{eqnarray*}

For both the regular and bilingual Moore-Lewis methods, data selection is performed by sorting the sentences according to the corresponding criterion and picking the top $n$ sentences (or sentence pairs). Determining the optimal value of $n$ is typically done empirically, training systems on subsets of increasing size, and evaluating on a held-out set.

\subsection{Class-based Language Difference Models for Data Selection}

The cross-entropy difference method can be improved by using language difference models (LDMs) instead of normal language models to compute the cross-entropy scores~\cite{AxelrodVyasMC:2015}. The standard and bilingual Moore-Lewis data selection methods use $n$-gram language models to calculate the cross-entropy difference scores needed to rank sentences in the data pool. However,  this creates a structural 
mismatch in the algorithm. The standard language models used in the computation are \textit{generative} models; they can be used to predict the next word. Yet, the actual cross-entropy difference score is \textit{discriminative} in nature, because it asks: is the sentence more like the task corpus, \underline{or} more like the pool corpus? 

This conceptual gap is well-known, and has led to data selection approaches that use classifiers to determine domain membership. However, to build a classifier is to fall into a trap! Only the Moore-Lewis \underline{score} is discriminative; the underlying corpora themselves are not. This is readily seen by noting that a sentence can appear in both the task and the pool corpora without any contradiction: ``task-ness" and ``pool-ness" are defined by construction rather than by any inherent characteristic. The two could overlap by 1\%, or by 99\%, and they would still be two corpora labeled `task' and `pool'.

It may help to reframe the `task' corpus as ``a pile of data that we already know we like'', and the `pool' corpus as ``a pile of data about which we do not yet have an opinion''. It is not necessary to know \underline{why} we like the data in the task corpus; it is also not necessary to have any opinion about whether the pool data looks useful, or not. With this view, the two corpora are not in opposition. Because they do not form opposing ends of a spectrum, then there is no underlying ``in-domain vs out-of-domain'' classification problem.\footnote{ We still use `in-domain' and `task' interchangeably.}

We previously defined a discriminative representation of the corpus as one that explicitly marks how the corpora differ. This helps quantify the difference between the task and the pool corpora. In~\cite{AxelrodVyasMC:2015}, every word in the corpora was replaced by a synthetic tag consisting of a class label and a discriminative marker. This procedure led to a representation of the text that explicitly encoded language differences between the corpora. 
Once the text had been transformed, the regular Moore-Lewis cross-entropy difference method is applied: two ``language models" are trained on the task and the pool. As the representation is discriminative, we have snuck discriminative information into the generative framework of the language models, so the two models are actually language difference models. Each sentence is then scored with the two models, and the scores are subtracted and used to sort the data pool and select the top $n$ lines. The bilingual version of class-based language difference models is exactly the same as bilingual Moore-Lewis: the corpus representation has changed, but the algorithm has not.

The tags in that work combined part-of-speech (POS) tags plus a suffix indicating the relative bias of each word. Both they and~\cite{KaziSaleskyTTGAEHOYH:2016} showed improved translation results when using the class-based difference labels to train the models for cross-entropy difference computation, instead of just using the words themselves. A variationused 20 class labels derived from an unsupervised POS tagger to create the language difference model~\cite{PhamNiehuesHCSW:2017} , but they did not obtain positive results when selecting monolingual data for back-translation and then subsequently using the artificially-parallel data to train a neural MT system. 


\section{Proposed Methods}   
\subsection{Cluster-Based Language Difference Models}

Using POS tags, as the basis for the discriminative tags that reduce the lexicon, creates a dependency on such a part-of-speech tagger. Such a tool is not always reliable, nor even available, for many languages nor specialized kinds of language. \cite{AxelrodVyasMC:2015}~posited that other methods of creating classes that capture underlying relationships within sentences (such as clustering or topic labels) might yield similar improvements. 

Following that hypothesis, we experimented with data selection using class-based language difference models. The synthetic difference representations were created using Brown cluster labels (generated from all of the words in the corpora) plus a relative-bias qualifier. Brown clustering~\cite{BrownDellapietraDLM:1992} is a way of partitioning a lexicon into classes according to the context in which the words occur in a corpus. Context, in this case, means the distribution of the words to their immediate left and right. 
The process of creating the clusters also generates a hierarchy above them, in the form of an unbalanced binary tree. Each word is assigned a bit string, and words that are statistically similar with respect to their neighbors will have similar bit strings and thus will be close together in the tree. An advantage of this method is that the number of clusters is freely specifiable, with a theoretical maximum of $V$, the size of the vocabulary. Choosing the correct amount is important, as too low a number would lead to poor-quality clusters, but generating a high number of them is computationally expensive.   

Following standard practice, we chose 1,000 as the number of clusters and added a suffix to indicate how much more 
likely a word is to appear in the task than in the pool corpus. Consistent with Table~1 
of~\cite{AxelrodVyasMC:2015}, we binned the probability ratios by order of magnitude (powers of $e$), 
from $e^{-3}$ to $e^3$. We indicated $e^{-3} < x < e^{-2}$ with the suffix ``\texttt{--}", $e^{1} < x < e^{2}$ as ``\texttt{+}", and so on. The following is an example of the text's new discriminative representation:
\noindent \begin{center} \vspace{-5mm}
\begin{tabular}{ll}
Original & {\em massive biotische krisen ... in} \\
& {\em vulkanen , gletschern , ozeanen .} \\
\hline
Transformed & {\tt 682/0 UNK/+ 935/0 3/- 7/0} \\
& {\tt 890/0 1/0 890/0 1/0 862/+ 2/0} \\
\end{tabular} \end {center}

The number before the slash indicates the cluster ID, and the marker after it represents the first digit of the log ({\em i.e.} exponent) of the ratio of the word's probability to appear in the task corpus divided by its probability in the pool corpus. 



Our class-based language difference model representation 
condensed the vocabulary of each of the corpora by at least 97\%. Table~\ref{table_cluster_vocab} contains the sizes of the corpus vocabularies before and after the cluster-based reduction.
\begin{table}[h]
\begin{center}
\begin{tabular}{lrrr}
Corpus & Word Types & Condensed & Reduction \\
\hline
Task (DE) & 93,767 & 1,691 & -98.20 \% \\
Task (EN) & 53,284 & 1,562 & -97.07 \% \\
Pool (DE) & 1,135,226 & 2,570 & -99.77 \% \\
Pool (EN) & 894,270 & 2,375 & -99.73 \% \\
\hline
\end{tabular}
\caption{\label{table_cluster_vocab} Size of the vocabularies 
that form the representation of the corpora used by the language difference models.}
\end{center}
\end{table}

It is on this transformed text that the language difference models were trained and the
cross-entropy difference scores computed. After ranking and selecting, the
sentences were transformed back to the original words and the MT systems were trained as usual.

\subsection{Cynical Data Selection}

The Moore-Lewis cross-entropy difference method has proved enduring, despite the subsequent development of several other methods with slightly better performance. 
Cross-entropy difference has had the advantage of being intuitive, reasonably effective, and easy to implement and integrate into existing MT pipelines. That said, it also has some structural problems.

Its subtractive relevance score implicitly defines the task and pool corpora as being opposing ends of a single spectrum: if the in-domain LM likes a sentence, it must be good, and if the pool LM likes it, then the sentence is irrelevant. This is never true, because language does not decompose cleanly into disjoint subsets, much less disjoint domains nor topics. The cross-entropy difference method is particularly weak when the task and pool corpora are similar, because the scores cancel out. Furthermore, the cross-entropy difference score indicates onlythat the selected sentences are well-liked by the in-domain model. It does not know whether the sentences are known to actually help model the in-domain data, nor if they even cover the in-domain vocabulary.

\textit{Cynical data selection}~\cite{Axelrod:2017} is a recent method to incrementally construct an efficient training corpus that models the in-domain corpus as closely as possible. Each sentence is scored by how much it would help model a particular task corpus, if it were added to the existing training corpus at the current iteration. The core idea was described as ``an incremental greedy selection scheme based on relative entropy, which selects a sentence if adding it to the already selected set of sentences reduces the relative entropy with respect to the in-domain data distribution''~\cite{SethyGeorgiouN:2006}. 

Cynical selection\footnote{ Sentences are only selected if they are of provable utility, regardless of whether an in-domain LM would like it, hence the name. } is an iterative algorithm that keeps track of how well the currently-selected data can model the task data. This is done via measuring the perplexity of a unigram LM trained on the selected sentences and evaluated on the in-domain corpus. The method iterates through all the words in the lexicon, and computes the expected entropy gain from adding a single instance of that word to the selected data. This step enables the algorithm to depend on the number of words in the lexicon rather than the number of sentences in the pool. The best word (that lowers entropy the most) is chosen. Given that word, the algorithm iterates through all the available (un-picked) sentences containing that word, and computes the expected entropy change from adding that single sentence by itself to the previously-selected set. The sentence with the most negative change is added, and the task perplexity is recomputed, taking into account the sentence that was just selected.

\section{Experimental Setup}

\subsection{Data}

We experimented on the German-to-English parallel data from the MT evaluation 
campaign for IWSLT 2017\footnote{\url{https://sites.google.com/site/iwsltevaluation2017/data-provided}}.
Our task data was the TED Talks corpus~\cite{CettoloGirardiF:2012}, 
comprising 218k parallel training sentences. The pool of available data consisted of 17.6M 
parallel sentences assembled from multiple sources: the preprocessed dataset from the WMT 2017 translation 
task\footnote{\url{http://www.statmt.org/wmt17/translation-task.html}} (containing 
the Europarl v7, Common Crawl and News Commentary corpora) and the OpenSubtitles2016 
collection\footnote{\url{http://opus.lingfil.uu.se/OpenSubtitles2016.php}}. 
We tuned on \texttt{dev2010} and tested on the concatenation of the \texttt{test2010, test2011, test2012, test2013, test2014}, and \texttt{test2015} datasets released for IWSLT 2017.

All corpora were preprocessed with the standard Moses~\cite{KoehnEtAl:2007} tools following the same pipeline 
employed in the preparation of the WMT 2017 preprocessed MT data\footnote{\url{http://data.statmt.org/wmt17/translation-task/preprocessed/de-en/prepare.sh}}. The sizes of the resulting datasets are in Table~\ref{table_data}.


\begin{table}[h]
\begin{center}
\begin{tabular}{llrr}
Corpus & Contents & Sentences & Tokens (DE)  \\
\hline  
Task & TED Talks & 218,020 & 4.0 M \\
\hline
Pool & WMT17 & 5,852,458 & 134.8 M \\
& OpenSubtitles 2016 & 11,811,574 & 100.1 M \\
& Total & 17,664,032 & 235 M \\
\hline
\hline
Tune & \texttt{dev2010} & 920 & 19.3 k \\
\hline
Test & \texttt{test2010-2015} & 8,431 & 154.8 k \\
\hline
\end{tabular}
\end{center}
\caption{\label{table_data} German-English parallel data statistics.}
\end{table}


\subsection{SMT Training Pipeline}

We trained our models with a statistical machine translation pipeline built upon
a combination of open-source tools. Input data was further subjected to various
normalizations, such as lowercasing, diacritic normalization, and the standardization of quotation marks. We split compound nouns on the German input using {\tt ASVToolBoox}~\cite{DBLP:conf/lrec/BiemannQHH08}. 

Translation was done with the {\tt Joshua} 
decoder~\cite{Ganitkevitch:2012:JPP:2393015.2393054}, an implementation of hierarchical phrase-based statistical machine translation. In some experiments, an additional target-side background language model was used while decoding, to promote fluent output and provide a more realistic use case. 
Word 
alignments were learned using {\tt fast\_align}~\cite{DBLP:conf/naacl/DyerCS13} 
with alignment models estimated in both directions and symmetrized using
{\em grow-diag-final-and}~\cite{OchNey:2003}. Grammar extraction was performed using the open-source framework 
{\tt Thrax}\footnote{\url{https://github.com/joshua-decoder/thrax}} and run on potent Elastic MapReduce (EMR) clusters during training. 
Tuning was done with the Margin Infused Relaxed Algorithm (MIRA)~\cite{DBLP:conf/emnlp/ChiangMR08} and optimized on BLEU~\cite{PapineniRoukosWZ:2002}. 

\subsection{Data Selection Tools}

The standard Moore-Lewis method uses $n$-gram language models to compute the cross-entropy score of each sentence according to the task and pool LMs. We used {\tt kenlm}~\cite{Heafield:2011} to estimate 6-gram Kneser-Ney (KN) smoothed language models, padding the vocabulary to 1.5M.

Our implementation of class-based difference models used Brown clusters and unigram frequency ratios to automatically produce discriminative representations of the task and pool corpora. These followed the steps in~\cite{AxelrodVyasMC:2015}, with every word replaced by a new token consisting of a cluster label and a bias suffix. 
 We employed the unsupervised Brown clustering algorithm~\cite{BrownDellapietraDLM:1992} for the construction of the clusters. The label+suffix tokens explicitly show how and where the two corpora's distributions differ from each other. We then trained 6-gram KN language models on this new representation. These models were used to compute cross-entropy difference scores over the new representation, and then the sentences were sorted by score. The discriminative representations were replaced by the original sentences after the selection process completed.

We wrote (and released\footnote{ \texttt{https://github.com/amittai/cynical} }) an open-source implementation of the cynical selection algorithm. Reducing the vocabulary size, by collapsing words into a single label, makes the algorithm's approximations tractable. We used the algorithm's default heuristics for vocabulary reduction, shown in Table~\ref{table_cynical_vocab}. Each criterion was applied (in order) to every word $v$ in the joint lexicon of the corpora. If the criterion was met, the word was replaced, and no further criteria were applied to the word. After all the criteria were applied, most of the vocabulary types had been collapsed down to a handful of labels, and the only words that remained intact were ones whose probability ratios were biased towards the task distribution.
\begin{table}[h]
\begin{center}
\begin{tabular}{lrl}
Criteria & Word Types & Replaced By  \\
\hline
$C_{\textsc{TASK}}(v) = 0$ & 1,050,590 & {\small \texttt{\_\_useless}}\\
$C_{\textsc{POOL}}(v) = 0$ & 9,131 & {\small \texttt{\_\_impossible}}\\
\vspace{-2mm} 
$C_{\textsc{TASK}}(v) < 3$  & 4,946 & {\small \texttt{\_\_dubious}}\\
\hspace{2mm} \textsc{and } $C_{\textsc{POOL}}(v) < 3$ & &\\
$\frac{P_{\textsc{TASK}}(v)}{P_{\textsc{POOL}}(v)} < e^{-1}$ & 3,945 & {\small \texttt{\_\_bad}}\\
$e^{-1} < \frac{P_{\textsc{TASK}}(v)}{P_{\textsc{POOL}}(v)} < e$ & 22,453& {\small \texttt{\_\_boring}}\\
\hline
\end{tabular}
\caption{\label{table_cynical_vocab} Criteria used to reduce the German lexicon from 1.14M to 29k for cynical data selection, baseed on the counts $C$ and probabilities $P$ for each word. Only the first criterion to match each word was applied, the word then being replaced by the corresponding tag.The second column shows how many word types were replaced by each rule. }
\end{center}
\end{table}

The intuition for the replacements is as follows: If a word $v$ does not appear in the task corpus, then it is \texttt{useless} for estimating relevance because it does not figure into the entropy calculations. If a word is in the task corpus but not in the pool, then it is \texttt{impossible} to change its empirical probability by adding sentences from the data pool. The probability of rare words (occurring once or twice in both corpora) cannot be estimated reliably, so their statistics are \texttt{dubious}. Words that are heavily skewed towards the pool distribution are \texttt{bad} for determining usefulness or information gain, because there is a danger that they will be over-represented in the selected sentences. We also tried appending a bias suffix to the \texttt{bad} label, following the procedure from the class-based language difference model approach. 

Even if we selected sentences randomly, we can expect to accurately estimate the probabilities of words occurring at roughly the same rate in both corpora, so their probability ratios are \texttt{boring}. We experimented with further dividing this category based on the frequency of these words in the task corpus, with the goal of limiting the number of sentences in which each token appears. This is important because the cynical algorithm implementation avoids computational complexity in terms of the number of sentences by replacing it with computational complexity in terms of the number of sentences in which words appear. However nearly every sentence in the pool contained a \texttt{boring} word, and it is not clear that this had any effect.

Due to the size of the pool corpus, we enabled the cynical data selection's ``batchmode'', where a variable amount ($\log k$) of sentences are selected per iteration. This variable batch size is computed from $k$, the number of sentences that contain the ``most useful word" for the current iteration.

\section{Experiments and Results}

We evaluated three data selection approaches: standard Moore-Lewis (monolingual and bilingual), class-based language difference models using Brown clusters (also monolingual and bilingual), and cynical data selection (monolingual only). The monolingual methods were used on each of the input and output languages, so we have results for all methods on both languages.
Our contribution is the first published use of Brown clusters for class-based language difference models, and also of cynical data selection. We present a head-to-head comparison of both, as well as comparing against the cross-entropy difference standards.

Each data selection method produced a subset of the pool corpus in which sentences are ranked by their relevance. The first four assign an absolute relevance score (some variation on cross-entropy difference) for each sentence. The cynical method provides a ranking, but the score for each sentence is the relative relevance score of each sentence with respect to all the higher-ranked sentences that precede it. For each experiment, we examined increasingly larger slices of the data ranging from the best $n = 100$k to the best $n = 12$M sentences out of the 17.6M sentence pairs available. 

\subsection{Perplexity of Modeling In-Domain Data}

\begin{figure}[h]
\centerline{\epsfig{figure=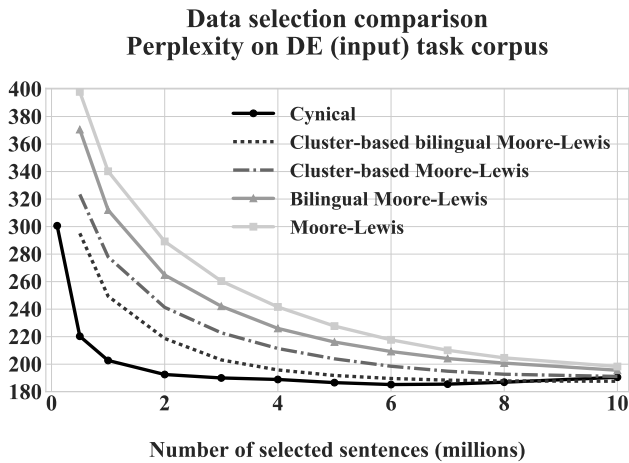,width=\columnwidth}}
\caption{Evaluating the selected data via perplexity scores on the TED DE (input) corpus. The cluster-based methods are better than their standard counterparts, and the cynical method is better still. }
\label{figure_ppl_de}
\end{figure}

\begin{figure}[h] 
\begin{center}
\epsfig{figure=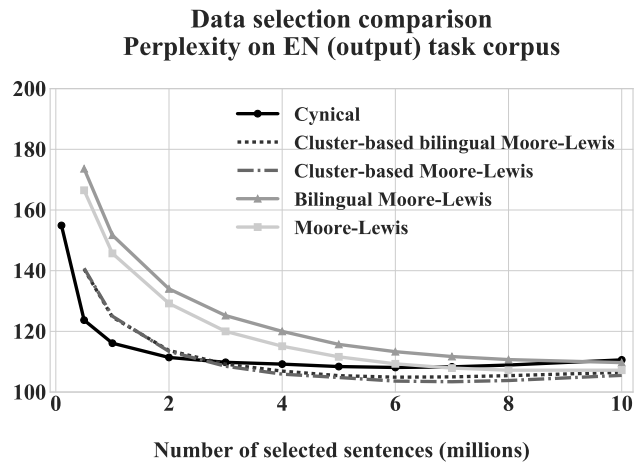,width=\columnwidth}
\caption{Comparison of perplexity scores on the TED EN (output) corpus. The new methods perform better than the standard approaches.}
\label{figure_ppl_en}
\end{center}
\end{figure}

In all cases, we first evaluated the selected data by itself, examining how well the selected data can model the task data. For this, we measured the perplexity and OOV count on the in-domain corpus, using models trained on only the selected data. For each of the data selection methods, we trained language models on the most relevant subsets of various sizes. The language models were similar to those used for selection ($n$-gram order 4, and vocabulary padded to 1.5M). We evaluated these models on their perplexity on the entire in-domain TED training set (218k sentences). Figure~\ref{figure_ppl_de} and~\ref{figure_ppl_en} show the full language modeling perplexity results for the input (German) and output (English) languages, respectively.

Each of the cluster-based data selection methods on the German side outperformed their vanilla Moore-Lewis counterparts (comparing monolingual cluster-based vs monolingual Moore-Lewis, both on the German side, and comparing bilingual cluster-based vs bilingual Moore-Lewis). At 6M sentences selected, near convergence, the cluster-based methods are each 20 perplexity points better than the standard cross-entropy difference, but the cynical selection method is slightly better. At 2M sentences selected, where the cynical method is nearly at its optimal perplexity, the gap between the cluster-based and standard approaches is 40 points, but the cynical method is 20 points better still, as highlighted in Table~\ref{table_lm_ppl_de}, despite being a monolingual method.

\begin{table}[h]
\begin{center}
\begin{tabular}{lrr}
Method & ppl at 2M & ppl at 6M\\
\hline
Moore-Lewis, mono (DE) & 289.2 & 217.7\\
Cluster-based, mono (DE) & 241.3 & 198.5\\ 
\hline
Moore-Lewis, bilingual & 264.8 & 209.2\\
Cluster-based, bilingual & 218.7 & 189.6\\
\hline
Cynical, mono (DE) & 192.5 & 185.2\\
\hline
\end{tabular}
\caption{\label{table_lm_ppl_de} Perplexity scores on the TED DE (input) corpus for the models trained with 2M and 6M selected sentences.}
\end{center}
\end{table}
				
On the English side, the improvements are similar in pattern though smaller in magnitude. This is expected, as English is easier to model than German. The cluster-based methods significantly outperform the regular Moore-Lewis methods. The cynical method once again converges the fastest of all the methods (after 2M sentences, compared to 6M for the others), though the cluster-based methods reach the lowest perplexity.

\subsection{Out-of-Vocabulary Rate on In-Domain Data}

\begin{figure}[h]
\centerline{\epsfig{figure=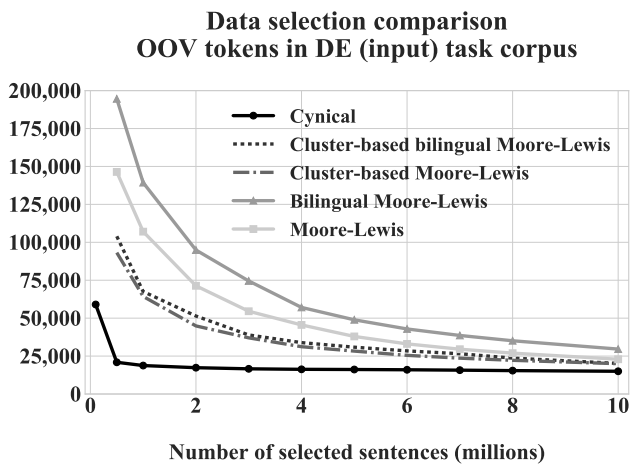,width=\columnwidth}}
\caption{Number of OOV tokens in the TED DE (input) corpus, according to LMs trained on the selected data.}
\label{figure_oov_de}
\end{figure}

\begin{figure}[h]
\centerline{\epsfig{figure=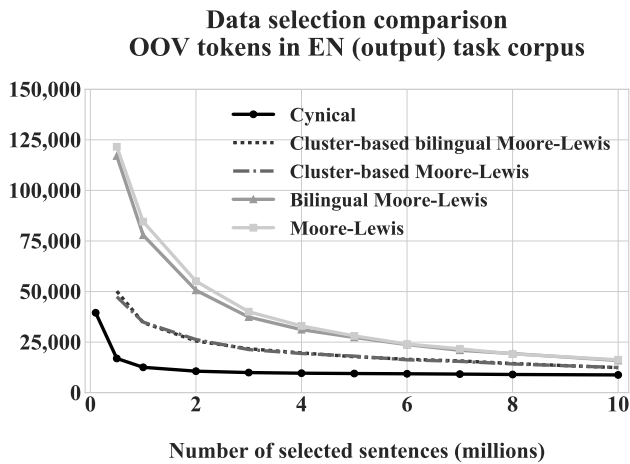,width=\columnwidth}}
\caption{Number of OOV tokens in the TED EN (output) corpus, according to LMs trained on the selected data.}
\label{figure_oov_en}
\end{figure}

Next, we computed the out of vocabulary (OOV) token count on the task corpus, using language models trained on only the selected data. Figures~\ref{figure_oov_de} and~\ref{figure_oov_en} show the OOV curves for the selected data with respect to the roughly 4M-token TED corpus in the input (German) and output (English) languages, respectively. In both graphs, the cluster-based Moore-Lewis methods converge to their final OOV count after selecting 6 to 8 million sentences. At the 6M sentence mark, the cluster-based methods have one-third fewer OOV tokens in the TED corpus than the vanilla Moore-Lewis methods. This substantial improvement corroborates the results from the method as proposed in~\cite{AxelrodVyasMC:2015}. 

However, the OOV rate of data selected using the new cynical data selection method is better still, by a large margin. At 1M sentences, the cynical subset has 85\% fewer OOVs than the monolingual Moore-Lewis, and 65\% fewer than the monolingual cluster-based version. More importantly, the first million sentences selected via the cynical method cover more of the task vocabulary than any quantity of data selected via the other methods. This rapid convergence to the minimum possible OOV rate -- the number of OOV tokens relative to all of the pool data -- results from the heuristic used by the cynical algorithm to select the word that needs to be covered by the next selected sentence. 

\subsection{Improving an In-Domain System with Selected Data}

Next, we performed an extrinsic evaluation, using the selected data to train machine translation models to be used in combination with the baseline in-domain system. In this way we tested the ability of the data selection methods to select subsets of the data that were actually useful in practice.

\begin{figure}[h]
\centerline{\epsfig{figure=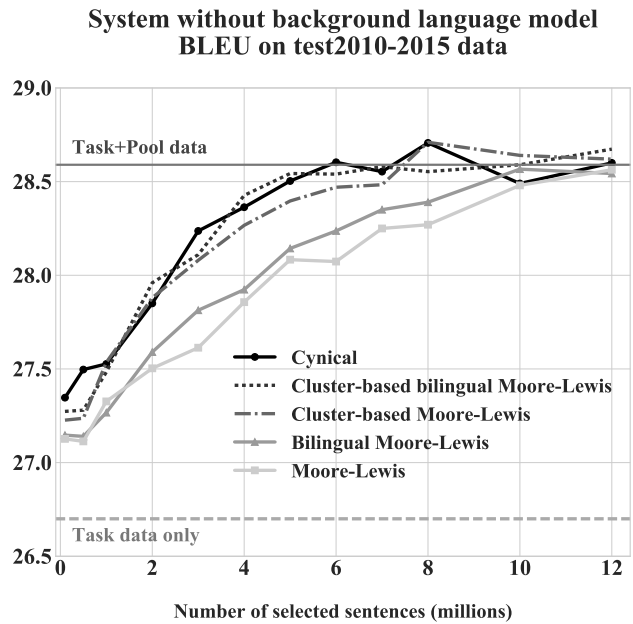,width=\columnwidth}}
\caption{Comparison of BLEU scores on multi-model systems using both task data and subcorpora (from 100k to 12M sentences) selected via each data selection method. The dashed line corresponds to a system trained on the task data only. The thin solid line indicates the result (28.59) obtained from training with the entire pool of 17.6M sentences.}  
\label{figure_bleu}
\end{figure}

Figure~\ref{figure_bleu} shows the machine translation results using BLEU. The horizontal dashed line is a static baseline that uses all of (and only) the available in-domain training data. The other curves are from multi-model systems where a model trained on selected data is used in combination with one trained on the task data. Each system curve in Figure~\ref{figure_bleu} shows the average score over 3 tuning and decoding runs, to mitigate the variability of MT tuning.

The baseline adapted systems using data selected via vanilla monolingual Moore-Lewis and the bilingual version performed better than the in-domain-only system, as expected. The difference between the monolingual and the bilingual versions' scores were minor, with the bilingual versions slightly ahead. The cluster-based versions of Moore-Lewis, which used language difference models to compute the cross-entropy difference scores, were roughly half a point better than the standard versions. The cynical methods performed as well as the fine-grained cluster-based approaches, despite collapsing approximately 850,000 vocabulary items down to a dozen coarse labels. 

BLEU scores corresponding to models trained with 2M and 6M selected sentences are compared in Table~\ref{table_bleu_model1}.
These results demonstrate that completely automatic clustering methods can be used to construct language difference models, so class-based version of cross-entropy difference need not depend on the availability of linguistically-derived labels. 

\begin{table}[h]
\begin{center}
\begin{tabular}{lrr}
Method & BLEU at 2M & BLEU at 6M\\
\hline
Moore-Lewis, mono (DE) & 27.50 & 28.07\\
Cluster-based, mono (DE) & 27.88 & 28.47\\
\hline
Moore-Lewis, bilingual & 27.59 & 28.24\\
Cluster-based, bilingual & 27.96 & 28.54\\
\hline
Cynical, mono (DE) & 27.85 & 28.60\\
\hline
\end{tabular}
\caption{\label{table_bleu_model1} BLEU scores on preprocessed data at 2M and 6M selected sentences from averaging 3 runs for a system, configured without background language model.}
\end{center}
\end{table}

\begin{figure}[h]
\centerline{\epsfig{figure=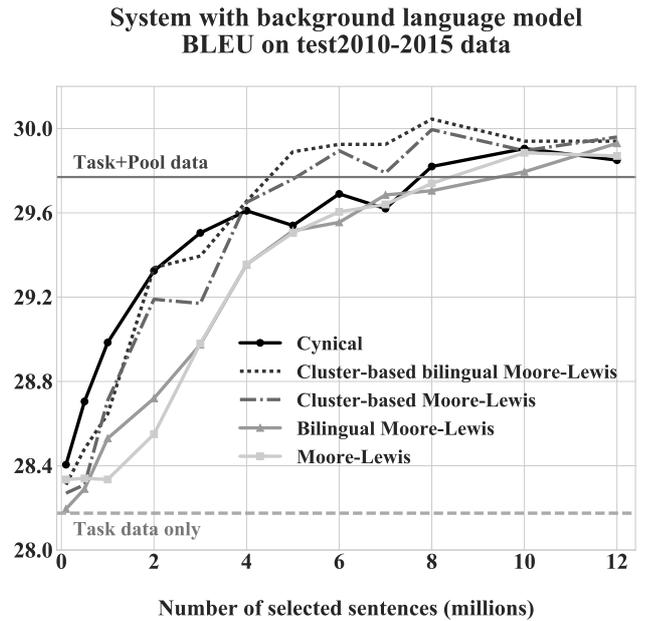,width=\columnwidth}}
\caption{Same as Figure~\ref{figure_bleu} but for the system with background language model. The BLEU score for the configuration with the entire pool data set is 29.77.}  
\label{figure_bleu_3LM}
\end{figure}

As a further test, we examined the use of these selected corpora inside a more robust system: one that has both in-domain parallel data, and a large background target-side language model. 
Figure~\ref{figure_bleu_3LM} shows the BLEU scores of multi-model systems that also incorporate a large background language model for decoding.

Table~\ref{table_bleu_model2} shows numeric values of the scores at 2M and 6M selected sentences, computed by averaging 2 training runs with target-side language model for fluency. Incorporating the background LM leads to overall score increases of +1 to +2 BLEU points compared to the results in Figure~\ref{figure_bleu}. Again, the cluster-based extension of Moore-Lewis outperforms the vanilla version. However, the cynical selection method exhibits bimodal performance: For smaller amounts of selected data, up to 4M sentences, it follows the performance of the class-based methods. After that, it switches sharply to tracking the performance of the vanilla methods. The gap is not large, so it might be due to jitter from tuning, but it is curious.

\begin{table}[h]
\begin{center}
\begin{tabular}{lrr}
Method & BLEU at 2M & BLEU at 6M\\
\hline
Moore-Lewis, mono (DE) & 28.55 & 29.61\\
Cluster-based, mono (DE) & 29.19 & 29.90\\
\hline
Moore-Lewis, bilingual & 28.72 & 29.56\\
Cluster-based, bilingual & 29.34 & 29.93\\
\hline
Cynical, mono (DE) & 29.33 & 29.69\\
\hline
\end{tabular}
\caption{\label{table_bleu_model2} BLEU scores on preprocessed data at 2M and 6M selected sentences from averaging 2 runs using the more robust configuration with background language model.}
\end{center}
\end{table}

\subsection{Better Matching of In-Domain Sentence Length}

\begin{figure}[h]
\centerline{\epsfig{figure=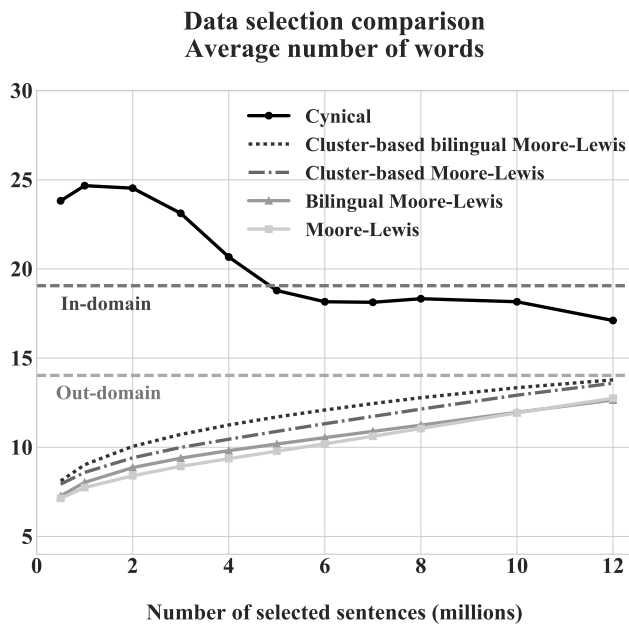,width=\columnwidth}}
\caption{Average number of tokens per sentence, as a function of selected corpus size.}
\label{figure_sentence_length}
\end{figure}

One of the advantages of data selection is that it allows for significantly smaller translation systems that perform at least as well as one trained on the full large-scale pool corpus. This holds true for all of the methods compared in this work: The size reduction of the translation systems is roughly proportional overall to the reduction in training corpus size. However, we noticed that the translation systems trained on the cynical subcorpora are twice as large as the ones selected using cross-entropy difference variants. We discovered that the Moore-Lewis style selection methods produced subcorpora that were almost identical in size, both on disk and in the number of tokens. The cynical method produced subsets containing significantly longer sentences than the other methods. Upon examination the sentences seemed fairly ordinary (\textit{i.e.} normal sentences, not particularly long), and it was the other methods that were producing significantly shorter sentences. Figure~\ref{figure_sentence_length} shows how the average sentence length changes with the number of sentences selected for each method.

The in-domain average sentence length is 19 tokens per sentence, and the pool corpus average is 14.
All of the Moore-Lewis variants have average sentence lengths even shorter than the pool average, and never greater. The cynical method mostly selected pool data matching the task average sentence length, despite having no explicit way to note the length of the selected sentences. This appears to substantiate the assertion that ``the length biases of the penalty and the gain terms counteract each other,  guarding the algorithm from the Moore-Lewis method's fixation on one-word sentences with a very common token"~\cite{Axelrod:2017}.

\section{Conclusions}

We have shown that both cluster-based language difference models and cynical data selection can be used to train better task-specific machine translation systems and more closely model a task corpus. This is the first published use of both methods.
Using Brown clusters instead of POS tags makes the language difference model variant of Moore-Lewis be both language- and situation-agnostic. We have not compared the two directly, but have shown that automatic clustering can be used successfully. This is good for domain adaptation scenarios where the particular kind of language is either low-resource or wildly different from the kind of data used to train NLP tools.

Also, we have presented empirical validation of the cynical selection method. Despite some of the crude algorithmic choices (4 labels for 97\% of the lexicon) as well as running in batch mode, the cynical selection method's performance matches the best variant of Moore-Lewis. Further improvements might well be possible with more fine-grained labels (e.g. adopting the Brown clustering labels from this work). The cynical method, as implemented, converges after roughly 66\% less data has been selected, compared to any of the cluster-based and vanilla Moore-Lewis methods, and has the best out-of-vocabulary word coverage. 

The tradeoff is that while cynical selection picks better sentences, leading to smaller selected corpora, it also uses significant amounts of RAM (60gb for 17m sentences) and time (1 day; after all, $n \log n$ is still super-linear in complexity). Our implementation is inefficient, but the memory requirements will always be larger than the class-based language difference model version of Moore-Lewis which was developed to reduce the run-time requirements for data selection. This is because cynical selection must store the entire pool corpus in memory, whereas the reduced lexicon of the class-based approach means the algorithm runs in roughly constant space. Where time or computation are at a premium, the cluster-based version is the best and most efficient version of cross-entropy difference. Where the resources are available, the cynical selection method is more accurate.

\section{Acknowledgements}
We wish to thank Stephan Walter, the Machine Learning Engineering team in Berlin, and Felix Hieber for their support and helpful discussions, as well as the anonymous reviewers for their detailed comments. We also wish to acknowledge the work of the Amazon Saar team to create the software infrastructure we used to train the MT systems. 

\bibliographystyle{IEEEtran}
\bibliography{references}

\begin{thebibliography}{10}
\providecommand{\url}[1]{#1}
\csname url@rmstyle\endcsname
\providecommand{\newblock}{\relax}
\providecommand{\bibinfo}[2]{#2}
\providecommand\BIBentrySTDinterwordspacing{\spaceskip=0pt\relax}
\providecommand\BIBentryALTinterwordstretchfactor{4}
\providecommand\BIBentryALTinterwordspacing{\spaceskip=\fontdimen2\font plus
\BIBentryALTinterwordstretchfactor\fontdimen3\font minus
  \fontdimen4\font\relax}
\providecommand\BIBforeignlanguage[2]{{%
\expandafter\ifx\csname l@#1\endcsname\relax
\typeout{** WARNING: IEEEtran.bst: No hyphenation pattern has been}%
\typeout{** loaded for the language `#1'. Using the pattern for}%
\typeout{** the default language instead.}%
\else
\language=\csname l@#1\endcsname
\fi
#2}}

\bibitem{AxelrodVyasMC:2015}
A.~Axelrod, Y.~Vyas, M.~Martindale, and M.~Carpuat, ``{Class-Based N-gram
  Language Difference Models for Data Selection},'' \emph{IWSLT (International
  Workshop on Spoken Language Translation)}, 2015.

\bibitem{Axelrod:2017}
A.~Axelrod, ``{Cynical Selection of Language Model Training Data},''
  \emph{arXiv [cs.CL]}, pp. 1--19, 2017.

\bibitem{MooreLewis:2010}
R.~C. Moore and W.~D. Lewis, ``{Intelligent Selection of Language Model
  Training Data},'' \emph{ACL (Association for Computational Linguistics)},
  2010.

\bibitem{AxelrodHeG:2011}
A.~Axelrod, X.~He, and J.~Gao, ``{Domain Adaptation Via Pseudo In-Domain Data
  Selection},'' \emph{EMNLP (Empirical Methods in Natural Language
  Processing)}, 2011.

\bibitem{KaziSaleskyTTGAEHOYH:2016}
M.~Kazi, E.~Salesky, B.~Thompson, J.~Taylor, J.~Gwinnup, T.~Anderson,
  G.~Erdmann, E.~Hansen, B.~Ore, K.~Young, and M.~Hutt, ``{The MITLL-AFRL IWSLT
  2016 Systems},'' \emph{Proceedings of the International Workshop on Spoken
  Language Translation (IWSLT)}, 2016.

\bibitem{PhamNiehuesHCSW:2017}
N.-Q. Pham, J.~Niehues, T.-L. Ha, E.~Cho, M.~Sperber, and A.~Waibel, ``{The
  Karlsruhe Institute of Technology Systems for the News Translation Task in
  WMT 2017},'' \emph{WMT Conference on Statistical Machine Translation}, 2017.

\bibitem{BrownDellapietraDLM:1992}
P.~F. Brown, V.~J. {Della Pietra}, P.~V. DeSouza, J.~C. Lai, and R.~L. Mercer,
  ``{Class-Based N-gram Models of Natural Language},'' \emph{Computational
  Linguistics}, vol.~18, no.~4, pp. 467--479, 1992.

\bibitem{SethyGeorgiouN:2006}
A.~Sethy, P.~G. Georgiou, and S.~Narayanan, ``{Text Data Acquisition for
  Domain-Specific Language Models},'' \emph{EMNLP (Empirical Methods in Natural
  Language Processing)}, 2006.

\bibitem{CettoloGirardiF:2012}
M.~Cettolo, C.~Girardi, and M.~Federico, ``{WIT{\^{}}3 : Web Inventory of
  Transcribed and Translated Talks},'' \emph{EAMT (European Association for
  Machine Translation)}, 2012.

\bibitem{KoehnEtAl:2007}
P.~Koehn, H.~Hoang, A.~Birch-Mayne, C.~Callison-Burch, M.~Federico,
  N.~Bertoldi, B.~Cowan, W.~Shen, C.~Moran, R.~Zens, C.~Dyer, O.~Bojar,
  A.~Constantin, and E.~Herbst, ``{Moses: Open Source Toolkit for Statistical
  Machine Translation},'' \emph{ACL (Association for Computational Linguistics)
  Interactive Poster and Demonstration Sessions}, 2007.

\bibitem{DBLP:conf/lrec/BiemannQHH08}
C.~Biemann, U.~Quasthoff, G.~Heyer, and F.~Holz, ``{ASV Toolbox: a Modular
  Collection of Language Exploration Tools},'' in \emph{LREC (Language
  Resources and Evaluation)}, 2008.

\bibitem{Ganitkevitch:2012:JPP:2393015.2393054}
J.~Ganitkevitch, Y.~Cao, J.~Weese, M.~Post, and C.~Callison-Burch, ``{Joshua
  4.0: Packing, PRO, and Paraphrases},'' in \emph{WMT (Workshop on Statistical
  Machine Translation)}, 2012.

\bibitem{DBLP:conf/naacl/DyerCS13}
C.~Dyer, V.~Chahuneau, and N.~A. Smith, ``{A Simple, Fast, and Effective
  Reparameterization of IBM Model 2},'' in \emph{NAACL (North American
  Association for Computational Linguistics)}, 2013.

\bibitem{OchNey:2003}
F.~J. Och and H.~Ney, ``{A Systematic Comparison of Various Statistical
  Alignment Models},'' \emph{Computational Linguistics}, vol.~29, no.~1, pp.
  19--51, mar 2003.

\bibitem{DBLP:conf/emnlp/ChiangMR08}
D.~Chiang, Y.~Marton, and P.~Resnik, ``{Online Large-Margin Training of
  Syntactic and Structural Translation Features},'' in \emph{EMNLP (Empirical
  Methods in Natural Language Processing)}, 2008.

\bibitem{PapineniRoukosWZ:2002}
K.~Papineni, S.~Roukos, T.~Ward, and W.-j. Zhu, ``{BLEU: A Method for Automatic
  Evaluation of Machine Translation},'' \emph{ACL (Association for
  Computational Linguistics)}, 2002.

\bibitem{Heafield:2011}
K.~Heafield, ``{KenLM : Faster and Smaller Language Model Queries},'' \emph{WMT
  (Workshop on Statistical Machine Translation)}, 2011.

\end{thebibliography}

\end{document}